\documentclass{egpubl-arxiv}
\usepackage{arxiv}

\SpecialIssueSubmission    

\usepackage[T1]{fontenc}
\usepackage{dfadobe}  

\usepackage{cite}  

\usepackage{amssymb}
\usepackage{pifont}
\usepackage{multirow} 
\usepackage{makecell}
\usepackage{gensymb}
\usepackage{footnote}
\usepackage{amsmath}
\usepackage{booktabs}
\usepackage{gensymb}
\usepackage{graphicx}

\newcommand{\eg}{\textit{e}.\textit{g}.}

\usepackage{xcolor}
\definecolor{darkred}{HTML}{CB4154}

\BibtexOrBiblatex
\electronicVersion
\PrintedOrElectronic

\usepackage{egweblnk} 

\title{ClotheDreamer: Text-Guided Garment Generation with 3D Gaussians}

\author[Y. Liu~et al.]
{\parbox{\textwidth}{\centering Yufei Liu $^{1}$ \orcid{0000-0003-0937-3340}
        Junshu Tang $^{2}$ \orcid{0000-0002-6549-5257}
        Chu Zheng $^{1}$ \orcid{0009-0006-8623-2620}
        Shijie Zhang $^{3}$ \orcid{0009-0007-7888-6947}
        Jinkun Hao $^{2}$ \orcid{0009-0005-1943-1307}
        Junwei Zhu $^{4}$ \orcid{0000-0002-5407-5150}
        Dongjin Huang $^{1}$ \thanks{Corresponding author.}
        }
        \\
{\parbox{\textwidth}{\centering $^1$Shanghai University, China
         $^2$ Shanghai Jiao Tong University, China \\
         $^3$ Fudan University, China 
         $^4$ Tencent Youtu Laboratory
       }
}
}

\usepackage{hyperref}

\begin{document}


\maketitle

\begin{figure*}[t]
  \centering
   \includegraphics[width=1.0\linewidth]{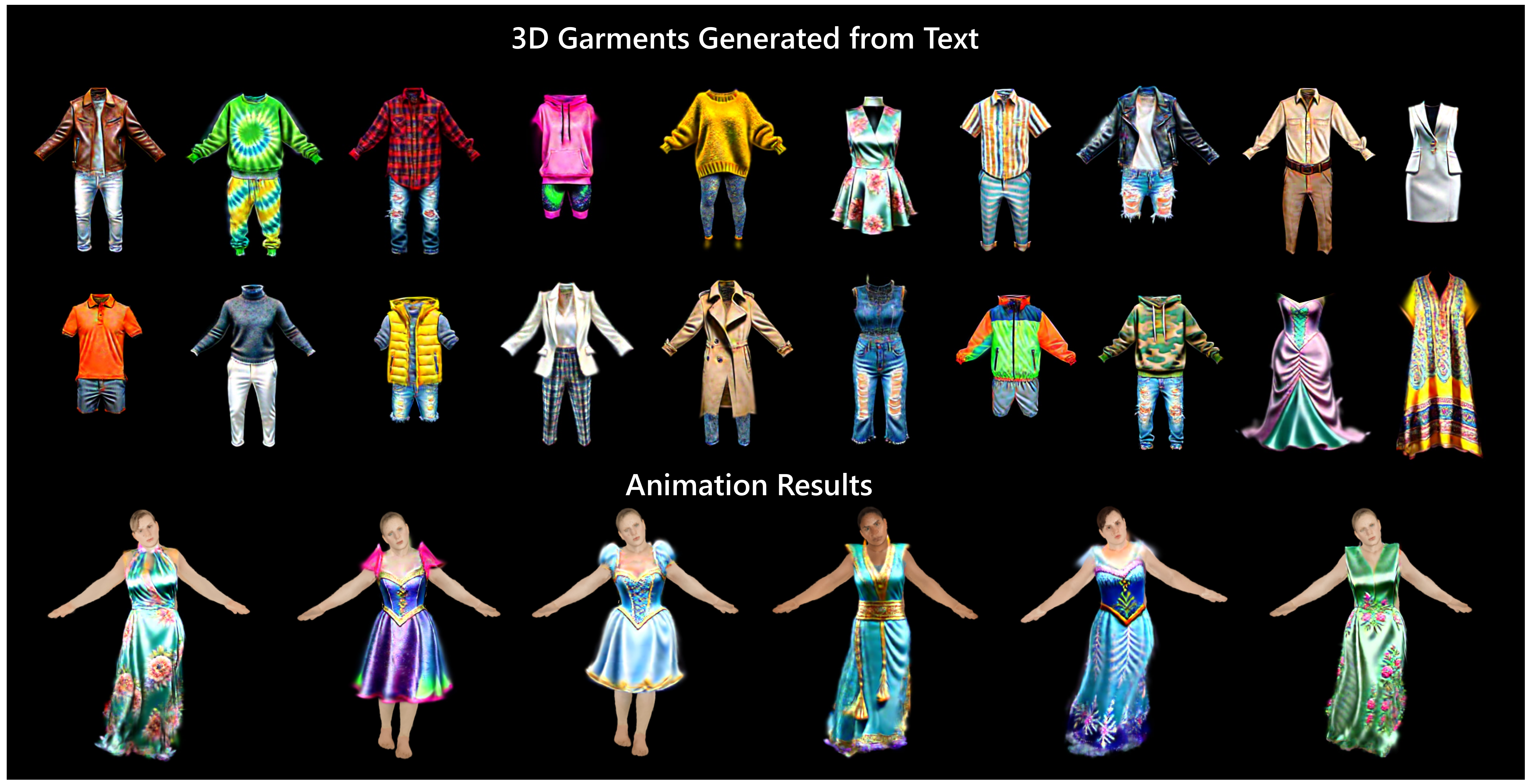}
   \caption{Overview of ClotheDreamer. We can generate diverse wearable production-ready 3D garment assets with refined geometry and realistic appearance details from text prompts, which enables accurate cloth animation.}
   \label{image: autofitting}
   \vspace{-1.5em}
\end{figure*}

\begin{abstract}
High-fidelity 3D garment synthesis from text is desirable yet challenging for digital avatar creation. Recent diffusion-based approaches via Score Distillation Sampling (SDS) have enabled new possibilities but either intricately couple with human body or struggle to reuse.
We introduce ClotheDreamer, a 3D Gaussian-based method for generating wearable, production-ready 3D garment assets from text prompts. We propose a novel representation Disentangled Clothe Gaussian Splatting (DCGS) to enable separate optimization. DCGS represents clothed avatar as one gaussian model but freezes body Gaussian splats.
To enhance quality and completeness, we incorporate bidirectional SDS to supervise clothed avatar and garment RGBD renderings respectively with pose conditions and propose a new pruning strategy for loose clothing. Our approach can also support custom clothing templates as input.
Benefiting from our design, the synthetic 3D garment can be easily applied to virtual try-on and support physically accurate animation. Extensive experiments showcase our method's superior and competitive performance.
Our project page is at \url{https://ggxxii.github.io/clothedreamer}.


\end{abstract}  

\section{Introduction}
\label{sec:intro}
3D Garment generation is of great need, as the application of fashion design, immersive interaction, and virtual try-on is becoming increasingly widespread. 
As a crucial part of 3D human creation, garment generation determines the realism and usability of the generated results. The procedure necessitates separate modeling of human and clothing components to enhance controllability.
3D garment creation in traditional industrial production involves a series of labor-intensive processes to create 2D sewing patterns and utilize physics-based simulations (PBS) to create realistic draping effects~\cite{provot1997collision,baraff2023large,narain2012adaptive,li2021deep}. A detailed garment may take several weeks and require meticulous manual design from artists. Recent diffusion-based 3D generation methods provide new opportunities for high-quality garment generation from text.

Traditional 3D content generation methods~\cite{Shap-e, point-e, get3d, magic123} aim to train generative models with large 3D datasets. Though it can quickly provide some reasonable 3D garment results, the clothing types are constrained within the training meshes. 
Recently, the advancement of diffusion-based optimization methods~\cite{fantasia3d, dreamgaussian, gaussiandreamer, lgm, magic3d, MVDiffusion, text2mesh} with Score Distillation Sampling (SDS) guidance proposed by DreamFusion~\cite{dreamfusion} shows promising 3D generation results from text. Mesh-based representation like DMTet~\cite{DMTET} is versatile and convenient for industrial pipeline integration.
However, without post-processing, it struggles to model the complex appearance of cloth material (\eg, yarn, lace, velvet, flannel). NeRF-based~\cite{nerf} ProlificDreamer~\cite{prolificdreamer} greatly improved realism with Variational Score Distillation (VSD) at the cost of time efficiency.
Utilizing 3D Gaussian Splatting (3DGS)~\cite{3dgs}, which combines the advantages of both mesh and NeRF-based representations,~\cite{gaussiandreamer, dreamgaussian, lgm} enhance 3D content generation quality with high training efficiency. However, when applied to garment generation, current approaches struggle to produce versatile reusable assets that accurately align with text descriptions. Text-guided clothed avatar generation methods~\cite{humangaussian, avatarclip, avatarcraft, avatarfusion, avatarbooth, tada} leverage SMPL body as mesh primitive to generate refined 3D avatars. However, the generated clothing is coupled with the human body, limiting additional manipulations such as style modifications, garment-swapping, or simulating realistic interaction between the clothing and the body, especially for loose clothing (\eg, long dress, gown). 

To tackle these challenges, we propose ClotheDreamer, an innovative 3D Gaussian-based approach for generating diverse and wearable 3D garments from text prompts. 
First, to separate clothing from the human body, we propose a novel Disentangled Clothe Gaussian Splatting (DCGS) representation, which enables separate optimization by initializing the clothed avatar as one-Gaussian model but freezes the body Gaussian splats. We discuss the importance of one-Gausssian model initialization in our methodology. Supporting robust clothing geometry generation, we initialize DCGS with language-based ID categorization derived from SMPL-X~\cite{smplx}.
Second, to efficiently regularize the underlying geometry of clothing Gaussians, we propose Bidirectional SDS guidance to supervise clothing and body Gaussian RGBD renderings with pose conditions. We also introduce a new pruning strategy for loose clothing to maintain the integrity of the garments. 
Finally, We further demonstrate our approach versatility by incorporating template mesh primitive for personalized generation. Besides rigid animation using Linear Blend Skinning (LBS)~\cite{smpl}, we can animate our DCGS garments with simulated mesh sequences. In addition, our generated garment also easily fits various body shapes by applying the transform matrix between input mesh vertices. 

In summary, our contributions are as follows: 
\begin{itemize}
\item We introduce ClotheDreamer, a novel 3D garment synthesis approach incorporating Disentangled Clothe Gaussian Splatting (DCGS). Our technique effectively decouples garments from the human body using ID-based initialization, facilitating the generation of diverse and wearable garment assets from text prompts.
\item We propose Bidirectional SDS guidance to efficiently regularize the underlying geometry of clothing Gaussians and a pruning strategy to enhance loose clothing completeness. 
\item Our approach supports customized clothing generation via template mesh guidance and enables accurate garment animation with simulated mesh prior. Extensive experiments demonstrate ClotheDreamer consistently outperforms existing methods in terms of text consistency and overall quality.
\end{itemize}

\section{Related Work} 
\label{sec:related}

\subsection{3D Garment Synthesis}
Traditionally, 3D garments are created with 2D sewing patterns and simulated through physics-based simulations (PBS) to achieve realistic draping effects~\cite{provot1997collision,baraff2023large,narain2012adaptive,li2021deep}, which involve meticulous manual design by artists, resulting in a laborous and time-consuming process. 
Consequently, many recent learned-based techniques~\cite{pergamo,wang2018learning,sizer,garnet,hood,nsf,lee2023multi} use neural networks to optimize this problem. The garments can be represented by 3D mesh templates~\cite{pbns,CAPE,VirtualBones,snug}, point clouds~\cite{deepsd,garnet++},UV maps~\cite{deepwrinkles,zhang2022motion,NSM,THuman3.0}, or implicit surfaces~\cite{dig,smplicit}. One common approach involves using linear skinning from SMPL~\cite{smpl}.
PBNS~\cite{pbns} and SNUG~\cite{snug} add physical constraints into self-supervision mechanisms for deep learning models, allowing training without actual garment data. However, they cannot manage meshes with varying topologies, even for the same garment, leading to the restriction of their applicability.
DrapeNet~\cite{drapenet} addresses this problem by using a single draping network conditioned on a latent code, enabling the generation and draping of previously unseen garments of any topology. 
To obtain better physical simulation results, DiffAvatar~\cite{diffavatar} proposes to utilize differentiable simulation for scene recovery and performs a strategy of body and garment co-optimization. Limited by the size and diversity of the training data, garments generated with traditional approach often exhibit fixed styles and patterns. 

\subsection{Text-Guided 3D Content Generation}
With the advancement of the diffusion models, recent works~\cite{fantasia3d, makeit3d, dreamgaussian, gaussiandreamer, lgm, prolificdreamer, magic3d, makeitvivid} have drawn considerable attention in the field of 3D content generation. ~\cite{Shap-e,point-e,get3d,holodiffusion,magic123} directly train 3D diffusion models to generate 3D representations. Leveraging Score Distillation Sampling (SDS) proposed by DreamFusion~\cite{dreamfusion}, we can optimize 3D scene renderings with pre-trained 2D generation model.  NeRF-based ProlificDreamer~\cite{prolificdreamer} greatly improved realism with Variational Score Distillation (VSD) loss, but suffers from long training time. Incorporating more efficient mesh-based representation,~\cite{fantasia3d} using DMTet~\cite{DMTET} with PBR materials to achieve realistic appearance modeling. DressCode~\cite{dresscode} introduces auto-regressive model with ChatGPT~\cite{chatgpt} to support customized sewing patterns for garment geometry generation from text. GarmentDreamer~\cite{garmentdreamer} adds normal and RGBA information into garment augmentation and employs implicit Neural Texture Fields (NeTF) combined with VSD to generate detailed texture. However, these mesh-based methods struggle to model hollow geometry or complex cloth materials with transparency (\eg, lace, wool), which traditionally requires layered PBR textures to represent. Implicit representation can capture refined appearance details.
Levering 3D Gaussian Splatting~\cite{3dgs}, recent works ~\cite{GSGEN,luciddreamer,dreamgaussian,gaussiandreamer,lgm} achieve more efficient rendering and higher quality generation. DreamGaussian~\cite{dreamgaussian} combines 3D Gaussians with SDS to generate 3D assets conditioned on a single image. GaussianDreamer~\cite{gaussiandreamer} improves 3D consistency by bridging the capabilities of 3D and 2D diffusion models, while LGM~\cite{lgm} presents a new mesh extraction method for 3D Gaussian by fusing information from multi-view images. Gaussian-based approaches allow for more flexibility in handling garment generation. However, current methods lack customization and face challenges in generating wearable garments with high-fidelity.  

\subsection{Text-Guided Clothed Avatar Generation}
Without strong mesh primitives, general 3D generation methods fall short in complex human-like contents. Many works~\cite{avatarclip, avatarbooth, avatarcraft, humangaussian, dreamavatar, dreamhuman, humannorm, avatarfusion, SO-SMPL} are dedicated to generating clothed avatars with SMPL/SMPL-X~\cite{smpl, smplx} as primitive. Representing with NeuS~\cite{neus}, AvatarCLIP~\cite{avatarclip} is the first to generate 3D avatars with CLIP~\cite{clip} guidance. AvatarCraft~\cite{avatarcraft}, DreamAvatar~\cite{dreamavatar}, and DreamHuman~\cite{dreamhuman} utilize diffusion guidance from large Text-to-Image (T2I) models to improve avatar quality. DreamWaltz~\cite{dreamwaltz} introduces a new 3D-consistent score distillation sampling to alleviate the Janus problem. 
Consequently,~\cite{text2mesh,tada, humannorm} explore usage with explicit representation for more convenient integration with modern graphics pipeline. TADA~\cite{tada} leverages deformable SMPL-X to generate animatable full-body avatar. HumanNorm~\cite{humannorm} and AvatarBooth~\cite{avatarbooth} fine-tune T2I models to customize avatar generation.
GAvatar~\cite{gavatar} and HumanGuassian~\cite{humangaussian} explore the application of 3DGS representation in generating 3D animatable clothed avatars with intricate details.
However, they typically represent the human body and the clothes as a holistic model, making it hard to produce reusable garment assets for further editing applications.~\cite{eva3d,LSVs} have attempted to create 3D avatars by separating components with layers, but the separation lacks physical significance.
To facilitate more convenient clothing modeling, similar to our intuitive idea is SO-SMPL~\cite{SO-SMPL}, which represents clothed avatars using separate meshes derived from SMPL. Recent concurrent work TELA~\cite{tela} uses NeRF to represent layered garments. Compared with these methods, our approach utilizes disentangled 3D Gaussians to produce wearable 3D garment assets with more intricate geometries and authentic appearances, which are also easy to animate.

\begin{figure*}[h]
  \centering
   \includegraphics[width=1.0\linewidth]{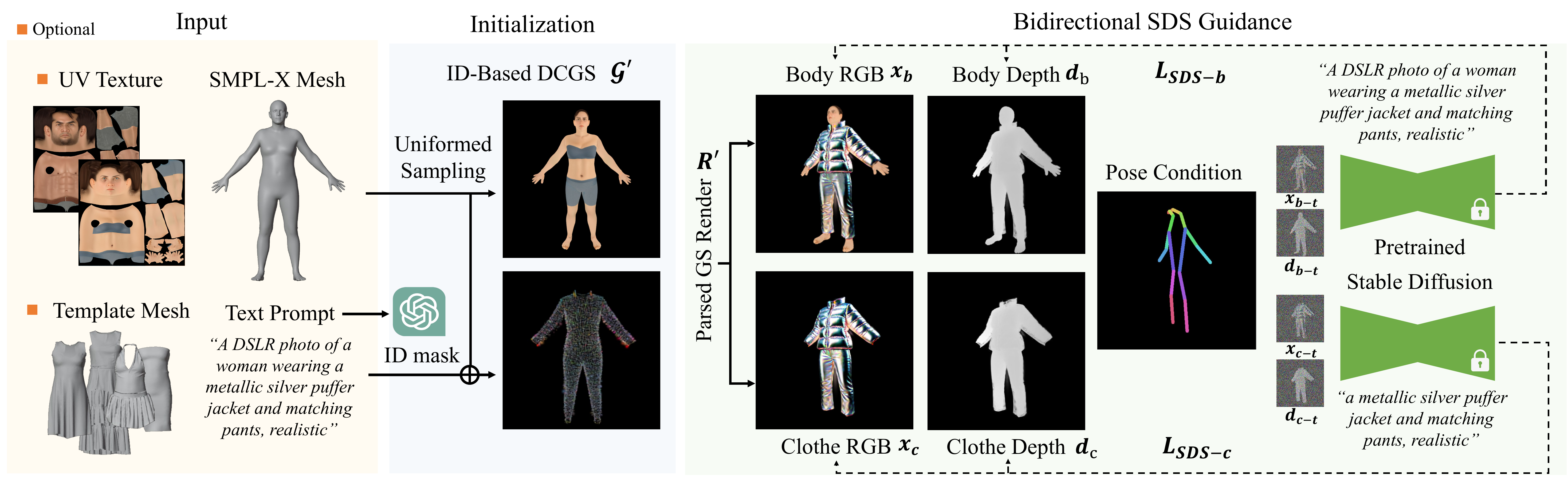}
   \caption{Overview of ClotheDreamer. Given a text description, we first leverage ChatGPT to determine clothing ID types for initialization. Our Disentangled Clothe Gaussian Splatting (DCGS) $\mathcal{G}^{'}$ represents clothed avatar as One-Gaussian model but freezes body Gaussian splats to achieve separate supervision. With parsed Gaussian Splatting (GS) render $\mathcal{R}^{'}$, we use Bidreactional SDS to guide clothing and body RGBD renderings separately with pose condition. We also support template mesh input for versatile personalized 3D garment generation.
   }
   \label{image: method}
   \vspace{-1.5em}
\end{figure*}

\section{Preliminaries} 
\label{sec:preliminaries}

\subsection{SMPL}
\label{subsec: smplx}
SMPL~\cite{smpl}, the Skinned Multi-Person Linear model, is a parametric human body model learning from extensive 3D scans using principal component analysis (PCA).
SMPL contains $N = 6890$ vertices and $K = 23$ joints (not including `pelvis'). It controls the shape and pose of the human body model by adjusting corresponding $\beta \in \mathbb{R}^{10}$ and $\theta \in \mathbb{R}^{24 \times 3}$ parameter in body template $\bar{\mathbf{T}}$. Human pose is represented by a kinematic joint tree, where the rotational relationship between each child node and parent node can be represented as $J(\vec{\beta})$. After applying linear blend skinning function $LBS$ to rest pose $T_P(\vec{\beta},\vec{\theta})$, the deformed vertices can be reposed by dual quaternion skinning. Then the SMPL model $M(\vec{\beta}, \vec{\theta})$ can be formulated as:
\begin{equation}
    \begin{aligned}
    \label{smpl_m}
    T_P(\vec{\beta}, \vec{\theta}) & = \bar{\mathbf{T}}+B_S(\vec{\beta})+B_P(\vec{\theta}), \\
    M(\vec{\beta}, \vec{\theta}) & = LBS\left(T_P(\vec{\beta}, \vec{\theta}), J(\vec{\beta}), \vec{\theta}, \mathcal{W}\right),
    \end{aligned}
\end{equation}
where $B_S(\vec{\beta})$ and $B_P(\vec{\theta})$ denote the offset from template. All SMPL models have a unified topology and consistent UV mapping. SMPL-X extends SMPL with fully articulated hands and an expressive face, providing a more accurate body model with 10475 vertices and 54 joints. With expression $\vec{\phi}$ and blend expression $B_E(\vec{\phi})$ , the SMPL-X can be represented as:
\begin{equation}
    \begin{aligned}
    T_P(\vec{\beta}, \vec{\theta}, \vec{\phi}) & = \bar{\mathbf{T}}+B_S(\vec{\beta})+B_P(\vec{\theta}) + B_E(\vec{\phi}), \\
    M(\vec{\beta}, \vec{\theta}, \vec{\phi}) & = LBS\left(T_P(\vec{\beta}, \vec{\theta}, \vec{\phi}), J(\vec{\beta}), \vec{\theta}, \mathcal{W}\right).
    \end{aligned}
\end{equation}

\subsection{3D Gaussian Splatting}
\label{subsec: 3dgs}
3D Gaussian Splatting (3DGS)~\cite{3dgs} serves as a hybrid approach for novel-view synthesis and 3D reconstruction, blending the strengths of both explicit and implicit radiance fields.
3DGS uses learnable 3D Gaussian $\mathcal{G}$ as a flexible and efficient representation of the underlying scene. Given a quired point location $p$, its center position $\mu \in \mathbb{R}^{3}$, and covariance $\Sigma \in \mathbb{R}^{7}$, one 3D Gaussian can be defined as:
\begin{equation}
    \mathcal{G}(p)=e^{-\frac{1}{2}(p-\mu)^T \Sigma^{-1}(p-\mu)},
\end{equation}
Each Gaussian is also associated with its opacity $\alpha \in \mathbb{R}$ and a set of spherical harmonics coordinates emitted by the Gaussian for all directions for color $c$ depiction:
\begin{equation}
    \mathbf{c}(p) =\sum_{i \in \mathcal{N}} c_{i} \sigma_{i} \prod_{j=1}^{i-1}\left(1-\sigma_{j}\right) ~~with~~\sigma_{i} =\alpha_{i} \mathcal{G}(p),
    \label{3dgs color}
\end{equation}
where $\sigma_{i}$ is the density of the $i$-th Gaussian. 
In addition, 3D Gaussian is like a blob and is centered at a point with a 3D covariance matrix $\Sigma$ in world space. To efficiently render and overcome the traditional gradient descent invalid issue, 3DGS using matrices for scaling $S \in \mathbb{R}^{3}$ and rotation $R \in \mathbb{R}^{3}$ expressed with 4 quaternions to parameterize $\Sigma$:
\begin{equation}
    \Sigma = RSS^{T}R^{T}
\end{equation} 
To handle 3D to 2D projection ambiguities, 3DGS optimization dynamically creates, alters, or removes geometry. Training process includes three main steps: 1. Structure from Motion (SfM)~\cite{sfm} Initialization, 2. Gradient Descent for Parameter Optimization, and 3. Adaptive Densification. All parameters can be optimized under the supervision of multiview images.
Thanks to the GPU-accelerated rasterization process, 3DGS can handle the projection process
with point-based rendering~\cite{point-based-rendering}, making it significantly faster than NeRFs. This combination of neural network-based optimization and explicit, structured data leads to improved quality, real-time, high-quality rendering with less training time, particularly for complex scenes and high-resolution outputs. We leverage 3D Gaussian to represent garments for easy editing or removal, higher render quality, and extending dynamic scenes with explicit time-step correspondences.


\subsection{Score Distillation Sampling}
\label{subsec: sds}
To lift 2D creation to 3D content generation, DreamFusion~\cite{dreamfusion} proposes Score Distillation Sampling (SDS), a loss function that samples from parameterized space. SDS utilizes 2D pre-trained large diffusion models as a prior to optimize 3D representation. When a 3D scene is parameterized by $\theta$ and rendered by a differentiable function $g(\cdot)$, the rendered image can be present as $x = g(\theta)$. Given a text prompt $y$, SDS optimizes 3D representation by minimizing the difference between the diffusion sampled noise $\epsilon_{\phi}$ and the noise $\epsilon$ added to the rendered image $x$. Thus the SDS loss is computed by:
\begin{align}
    \label{sds}
    \nabla_{\theta} \mathcal{L}_{S D S}=\mathbb{E}_{t, \epsilon}\left[w(t)\left(\epsilon_{\phi}\left(x_{t} ; y, t\right)-\epsilon\right) \frac{\partial x}{\partial \theta}\right],
\end{align}
where $t$ is the noise level, $x_t$  is the noisy version of $t$, $\hat{\epsilon}_{\phi}\left(x_{t}; y, t\right)$ is the predicted noise by diffusion and $w(t)$ is a weighting function control the noise schedule decided by the diffusion sampler. 

\section{Methodology}
\label{sec: method}
In this paper, we introduce a 3D Gaussian-based method for high-fidelity reusable 3D garment generation via text guidance, named ClotheDreamer. The overview of our framework is shown in Figure~\ref{image: method}.
In order to improve the interaction of the synthetic garment with the human body, we proposed a novel Disentangled Clothe Gaussian Splatting (DCGS) which separates the clothed body into the SMPL body part and the garment part. 
In Section~\ref{subsec: Zero-shot Garment Generation}, we begin to initialize DCGS with relative semantic ID from SMPL according to the text instructions.
Then we propose to learn the disentanglement between body parts and garments by manipulating the optimization gradient.
In Section~\ref{subsec: Bidirectional SDS guidance}, we introduce Bidirectional SDS guidance on individual renderings and a new pruning strategy for loose clothes. We further show the effectiveness of our framework which can achieve template-guided generation to ease customization in Section~\ref{subsec: Template-guided Garment Generation}. Finally, we show the process of animating the synthetic garment with diverse body motions in Section~\ref{subsec: Animate DCGS Garment}.

\subsection{Zero-shot Garment Generation}
\label{subsec: Zero-shot Garment Generation}

\textbf{ID-Based DCGS Initialization.} Human garments exhibit diverse and intricate shapes. Previous methods use SfM~\cite{sfm} points or generic point clouds generated by Shap-E~\cite{Shap-e} and Point-E~\cite{point-e} as initial points, which fall short in the human category and struggle to provide a strong prior. 
A recent popular initialization choice for clothed avatars is to use coherent body models like SMPL or SMPL-X, as they are well-structured and parameterizable for optimization. However, these methods typically sample points over the entire body mesh surface, making it difficult to manipulate clothing separately.

\begin{figure}[t]
  \centering
   \includegraphics[width=1.0\linewidth]{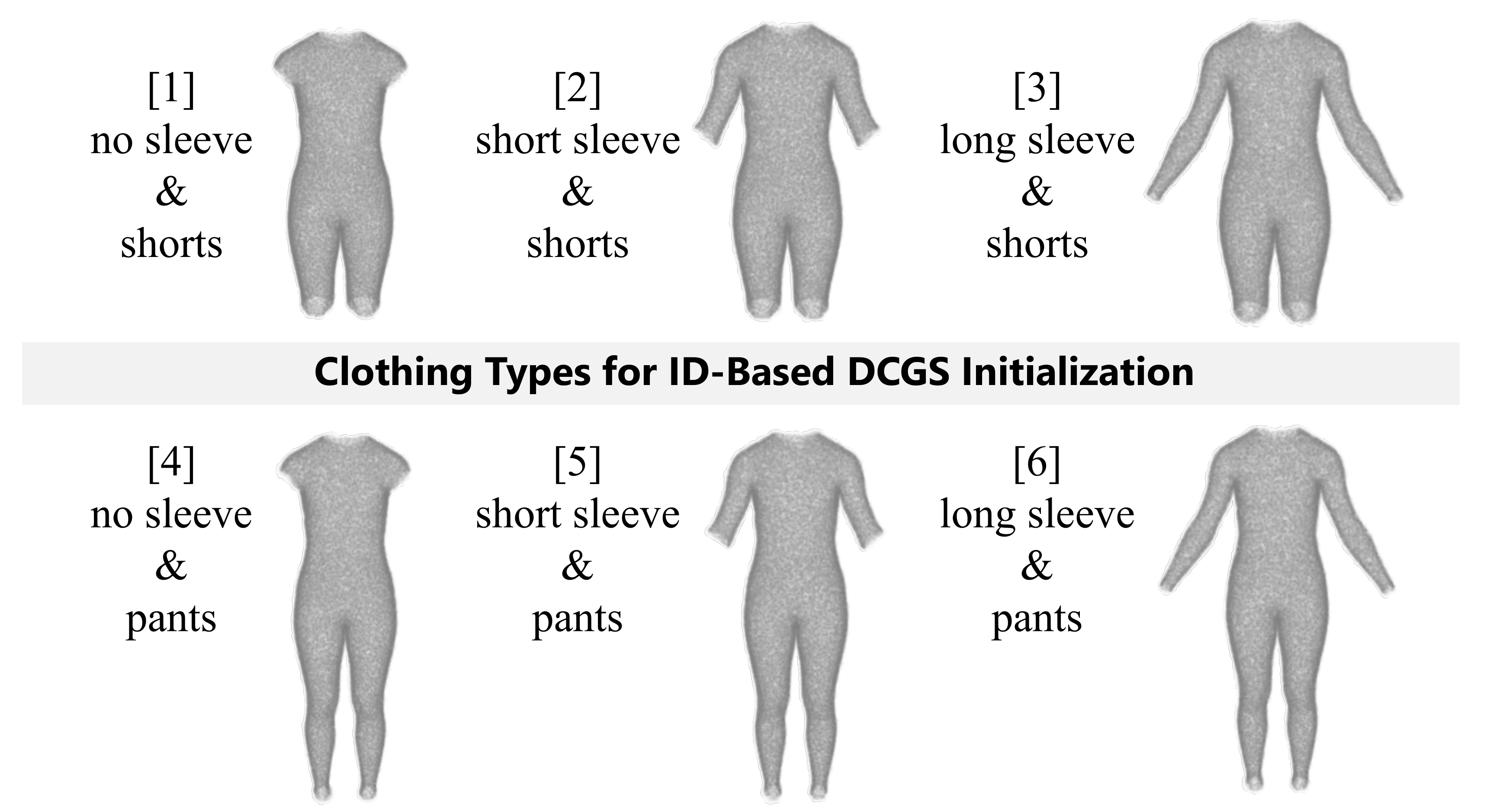}
   \caption{Clothing types. We offer six common groups for initializing DCGS in our zero-shot garment generation.}
   \label{image: ids}
   \vspace{-2em}
\end{figure}

We instead use an ID-based initialization. Besides uniform mesh topology, the SMPL model also provides body segmentation. By parsing joint binding, we can use vertex IDs to select corresponding body faces. We leverage SMPL-X 55 segmentation parts and offer six common groups for different clothing types. The visualization of each group is shown in Figure~\ref{image: ids}. We sample the clothing points proportional to the selected ID with the initial body points. Specifically, given one SMPL-X model with vertices $V_{b}$ and sample count $P_{b}$, and the clothing points $P_{c}$ is computed by:
\begin{equation}
    P_{c} = \frac{idx}{V_{b}} P_{b},
    \label{eq: points}
\end{equation}
where $idx$ is the predefined vertex set, $P_b=100K$. 
We sample all clothing points with united scaling, mean color, and no rotation, see Section~\ref{subsec: experimental setup} for details.
For precise supervision, body points can be initialized by utilizing body primitives on a 2D grid in the mesh's UV texture space and generating the 3D points on the mesh surface corresponding to the UV coordinates. This results in a textured body along with mean-colored clothing. We utilize base human textures generated with~\cite{texdreamer}. For a more efficient categorization process, we leverage GPT-4~\cite{chatgpt} to classify input descriptions based on common knowledge of clothing. Before cloth sampling, we scale and transform the SMPL-X mesh to the center of the 3D rendering space.
To generate high-quality clothing assets, we aim to optimize the clothing separately. Based on our DCGS representation, we can combine 3D Gaussians created from two point clouds for rendering but only optimize the clothing part. 

\begin{figure}[t]
  \centering
   \includegraphics[width=1.0\linewidth]{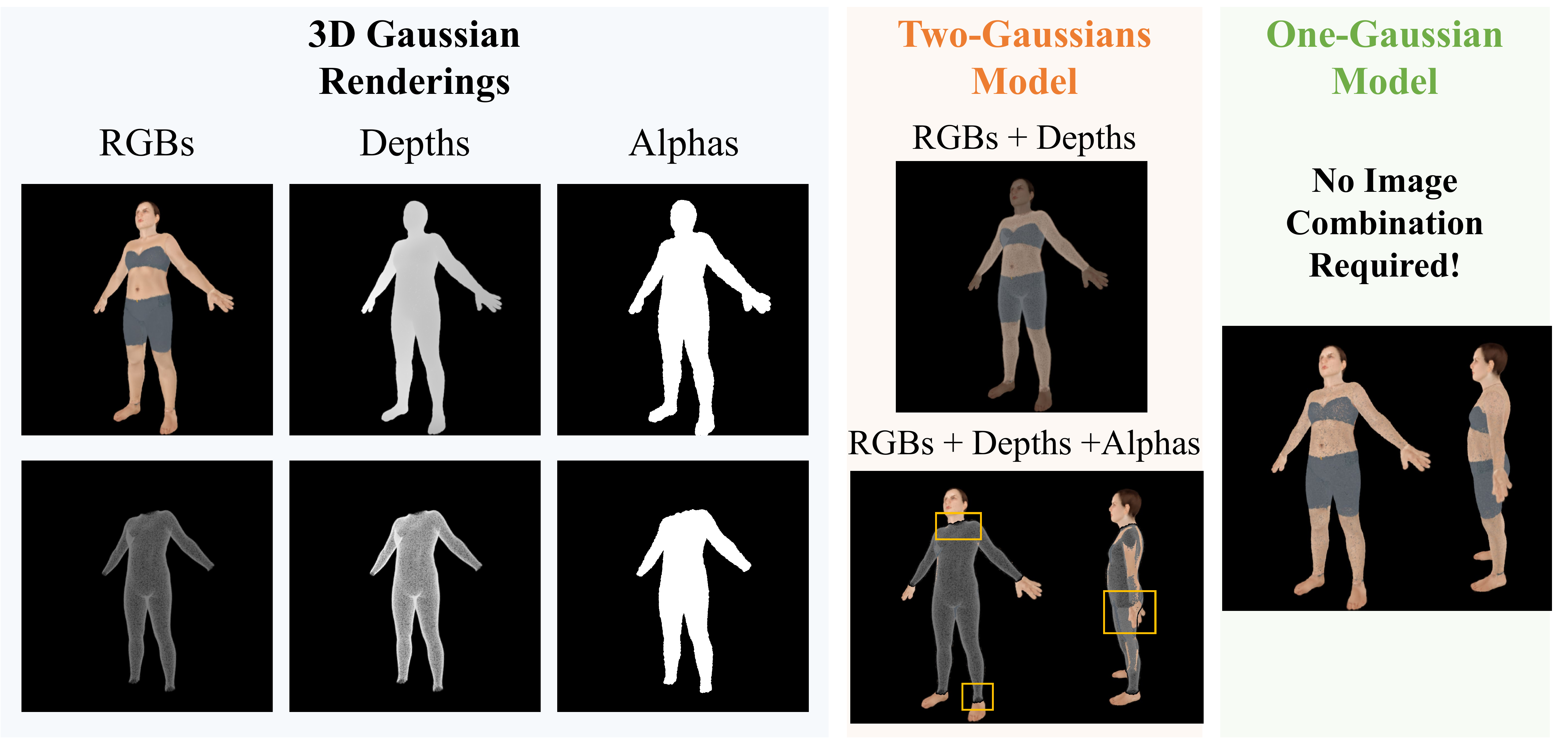}
   \caption{Importance of One-Gaussian Initialization. Artifacts may appear for clothed avatar rendering with Two-Gaussians model. Please kindly zoom in for a better comparison.}
   \label{image: 1gs2gs}
   \vspace{-2em}
\end{figure}

\noindent\textbf{Two-Gaussians Model.}
An intuitive approach is to create two Gaussian models to ensure correct gradient propagation and merge their RGB images using corresponding depth or alpha renderings. However, as shown in Figure~\ref{image: 1gs2gs}, this may cause artifacts for rendering clothed avatar.
The view-dependent depth map can be computed by accumulating depth value overlapping the pixel of $\mathcal{N}$ ordered Gaussian instances via point-based $\alpha$-blending:
\begin{equation}
    \mathbf{d}(p) =\sum_{i \in \mathcal{N}} d_{i} \sigma_{i} \prod_{j=1}^{i-1}\left(1-\sigma_{j}\right) ~~with~~\sigma_{i} =\alpha_{i} \mathcal{G}(p),
\end{equation}
where $ d_{i}$ is the depth of the $i$-th Gaussian and $\mathcal{G}(p)$ is the value of the $i$-th Gaussian at the queried point $p$ as defined in Equation~\ref{3dgs color}. Since the depth and opacity rendering are updated during optimization, Gaussian needs a fair amount of training steps to get correct renderings. Depths from a single step cannot accurately determine the occlusion relationships between the two objects. Combining body and clothe RGB rendering using depths will lead to partially transparent and unsatisfying results. Additionally applying alpha renderings as a pre-processing combined mask can cause uneven black edges due to the Gaussian point-based representation. Thus if we represent clothing and body as independent Gaussian models, it's impossible to fit the clothing onto the body for holistic optimization, which is essential for the initial fitting. 

\noindent\textbf{One-Gaussian Model.}
Therefore, we tend to initialize the body and clothing as one single Gaussian model, which can correctly render clothed avatar with separate point clouds, see Figure~\ref{image: 1gs2gs}.
We dive into 3D Gaussian back-propagation process and propose to mask the gradient to avoid optimization for selected body splats. We still use stochastic gradient descent to fit the scene properly by estimating Gaussian parameters. Positional updates follow exponential decay, guided by a loss function combining $\mathcal{L}_{1}$ and D-SSIM term:
\begin{equation}
    \mathcal{L}_{3dgs} =(1-\lambda) \mathcal{L}_{1}+\lambda \mathcal{L}_{\text {D-SSIM }}
\end{equation}

We keep track of the sampled body points and mask their gradients after every backpropagation process. During the adaptive densification process, we mask the Gaussian position gradient accumulation tensor to update with partial valid gradients before rasterization. Additionally, we only select clothing points for densification or pruning determination to reduce computational cost.

\subsection{Bidirectional SDS Guidance}
\label{subsec: Bidirectional SDS guidance}
Given that the SMPL-X prior merely acts as an initialization, more comprehensive guidance is required to enhance 3DGS training. We follow~\cite{humangaussian} and utilize the pretrained diffusion model to supervise both RGB and depth rendering. Specifically, to inject joint information, we replicate the diffusion UNet backbone to deal with the denoising of each target. We also leverage additional pose conditions via channel-wise concatenation. With the assistance of structural expert branches, the trained diffusion can simultaneously denoise the image RGB and depth. To increase diffusion spatial alignment ability, we also incorporate pose images extracted from SMPL-X joints as a condition besides input prompts. We parse GS render for garment $\mathcal{R}^{'}_{b}$ and clothe avatar $\mathcal{R}^{'}_{c}$ to provide separate renderings from one Gaussian model $\mathcal{G}$:
\begin{equation}
    \begin{aligned}
        x_{b}, d_{b} = \mathcal{R}^{'}_{b}(\mathcal{G}^{'}(P_{b})), \\
        x_{c}, d_{c} = \mathcal{R}^{'}_{c}(\mathcal{G}^{'}(P_{c})), 
    \end{aligned}
\end{equation}
where, respectively for the body and clothing, $x_{b}$ and $x_{c}$ are RGB renderings, $d_{b}$ and $d_{c}$ are depth renderings, and $P_{b}$ and $P_{c}$ are queried Gaussian points.
During DCGS optimization, we simultaneously employ two bidirectional SDS to supervise the garment and avatar on their RGBD renderings. With the normalized clothing and body depth map, our loss can be computed by:
\begin{equation}
    \begin{aligned}
\nabla_{\theta} \mathcal{L}_{\mathrm{SDS-b}} & =\lambda_{1} \cdot \mathbb{E}_{\boldsymbol{\epsilon}_{\mathbf{x_{b}}}, t}\left[w_{t}\left(\boldsymbol{\epsilon}_{\phi}\left(\mathbf{x}_{b-t} ; {p_{b}}, y\right)-\boldsymbol{\epsilon}_{\mathbf{x_{b}}}\right) \frac{\partial \mathbf{x_{b}}}{\partial \theta}\right] \\
& +\lambda_{2} \cdot \mathbb{E}_{\boldsymbol{\epsilon}_{\mathbf{d_{b}}}, t}\left[w_{t}\left(\boldsymbol{\epsilon}_{\phi}\left(\mathbf{d}_{b-t} ; {p_{b}}, y\right)-\boldsymbol{\epsilon}_{\mathbf{d_{b}}}\right) \frac{\partial \mathbf{d_{b}}}{\partial \theta}\right],
    \end{aligned}
\end{equation}

\begin{equation}
    \begin{aligned}
\nabla_{\theta} \mathcal{L}_{\mathrm{SDS-c}} & =\lambda_{1} \cdot \mathbb{E}_{\boldsymbol{\epsilon}_{\mathbf{x_{c}}}, t}\left[w_{t}\left(\boldsymbol{\epsilon}_{\phi}\left(\mathbf{x}_{c-t} ; {p_{c}}, y\right)-\boldsymbol{\epsilon}_{\mathbf{x_{c}}}\right) \frac{\partial \mathbf{x_{c}}}{\partial \theta}\right] \\
& +\lambda_{2} \cdot \mathbb{E}_{\boldsymbol{\epsilon}_{\mathbf{d_{c}}}, t}\left[w_{t}\left(\boldsymbol{\epsilon}_{\phi}\left(\mathbf{d}_{c-t} ; {p_{c}}, y\right)-\boldsymbol{\epsilon}_{\mathbf{d_{c}}}\right) \frac{\partial \mathbf{d_{c}}}{\partial \theta}\right].
    \end{aligned}
\end{equation}
where $\lambda_{1}$ and $\lambda_{2}$ are coefficients that balance the effects between RGB and depth. We compute $\mathcal{L}_{\mathrm{SDS-b}}$ and $\mathcal{L}_{\mathrm{SDS-c}}$ according to our parsed random render $\mathcal{R}^{'}$ for either clothing or body supervision.

\begin{figure}[t]
  \centering
   \includegraphics[width=1.0\linewidth]{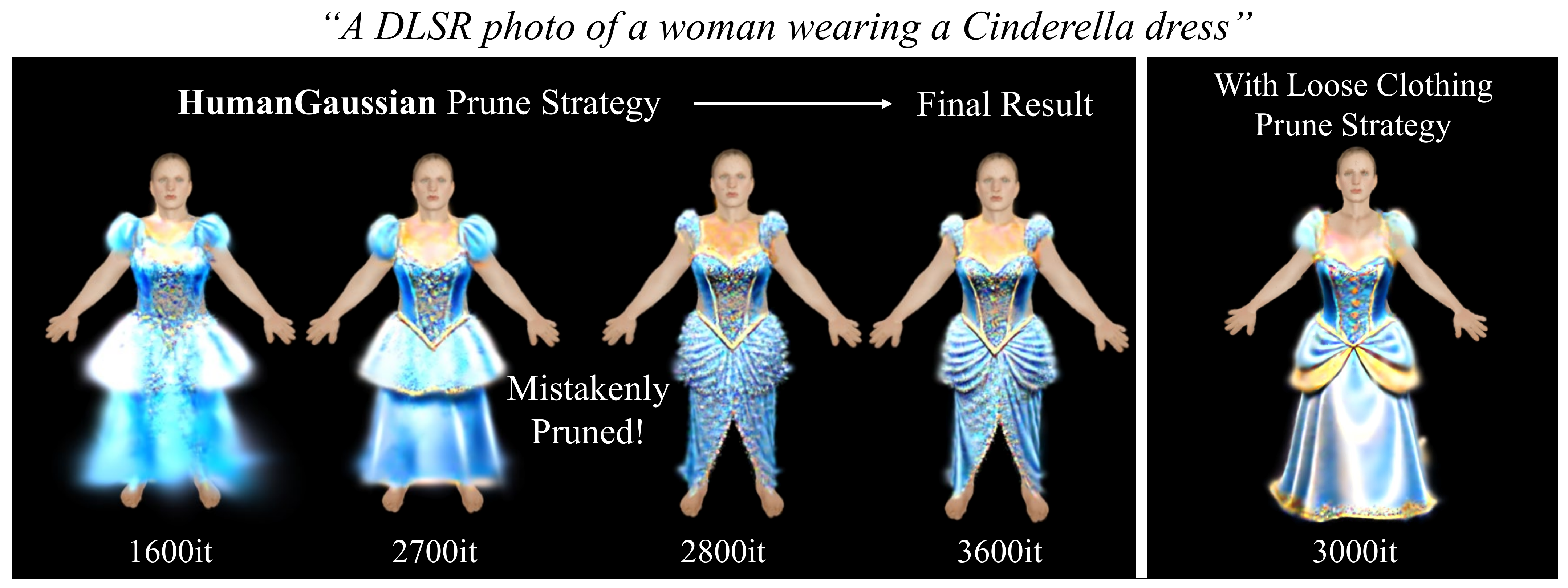}
   \caption{Improper prune strategy~\cite{humangaussian} may mistakenly eliminate useful Gaussian splats for loose clothing.}
   \label{image: prune}
   \vspace{-2em}
\end{figure}

We observed that the bidirectional SDS guidance can enhance the quality of the generated clothing and reduce Gaussian artifacts, see ablation in Section~\ref{subsec: Ablation Study}. 
However, the SDS supervision may be unstable and cause clothing geometry distortion.
~\cite{humangaussian} size-conditioned Gaussian pruning strategy in prune-only phase is effective in general but may mistakenly eliminate useful Gaussian points for loose clothing cases (\eg, long dresses, gowns). As shown in Figure~\ref{image: prune}, a reasonable result is obtained with Gaussian in the early optimization stage. However, due to improper pruning strategy, a large amount of useful Gaussians are pruned, resulting in an unsatisfactory generation. Therefore, we propose to use a new prune strategy for loose clothing generation. We reduce the training step and only prune one time during the middle stage of training, with an increased scaling factor range as the condition. This ensures the elimination of overly stretched points while preserving the completeness of the generated clothing. See Section~\ref{subsec: experimental setup} for details. 

\subsection{Template-Guided Garment Generation}
\label{subsec: Template-guided Garment Generation}
To increase practical use for personalization, we show a novel technique for guiding 3D garment generation with custom template garment mesh. The template mesh is meant to serve as a holistic shape guidance for garment generation. We aim to encourage Gaussians to densify near the input mesh shape in general but not limited by the surface position. The initiative way to achieve this is to change sampled clothing points during zero-shot DCGS initiation. However, custom template meshes may possess much different scale and position information. A naive changing of clothing sample points will result in misplaced 3D Gaussians. 

We therefore first move input template clothing and SMPLX mesh to their centers for location alignment in world space. Then we concatenate two sets of vertices for scaling and transforming unitedly in the local space. We keep track of each part and sample 3D Gaussian points separately after the transformation. Given a mesh-based primitive, we can generate diverse clothing geometry and texture details from text descriptions, while utilizing initial points to constrain the overall clothing style. Using clothes created with Marvelous Designer~\cite{MD}, we show some template-guided garment generation results in Figure~\ref{image: template}.

\subsection{Animate DCGS Garment}
\label{subsec: Animate DCGS Garment}
We demonstrate two possible ways to animate our DCGS garment asset. For tight-fitting clothing (\eg, shirts, short sleeves, jeans), we register the trained Gaussian avatar as SMPL-X and animate it using a sequence of SMPL-X pose parameters.
However, this process treats the generated Gaussians as one integrated model and struggles to properly animate loose clothes (\eg, gown, skirt, long dress). The lower clothing Gaussians are separately bound to SMPL-X legs, causing tearing artifacts during animation. With our representation, the generated garment is fully decoupled from the avatar body, we propose to effectively animate generated DCGS garments with simulated mesh prior.

\noindent\textbf{Mesh-Based Garment Animation.} Benefiting from Gaussian explicit representation, we can treat the generated Gaussian assets as a point cloud. We aim to leverage temporal simulated mesh as priors to drive the clothing points. Marvelous Designer is a powerful and advanced software for 3D virtual clothing simulation that leverages physics parameters. With a simulated garment mesh sequence, we can easily animate our DCGS asset with three steps: 1. Iterative Closest Point (ICP) Registration~\cite{icp}, 2. K-Dimensional Tree (KD-tree) Binding~\cite{kdtree}, and 3. Gaussian Transformation. We first use ICP registration for coarse alignment between the DCGS asset and the first the simulated mesh. For accelerated and refined nearest neighbor search, we construct a KD-tree to find the nearest mesh point for each clothing Gaussian. We utilize the deformation between mesh vertices to compute the transformations of clothing Gaussians. Note that our approach is capable of animating multiple similar-shaped garments using just one simulated mesh sequence.

\begin{figure*}[h]
  \centering
   \includegraphics[width=1.0\linewidth]{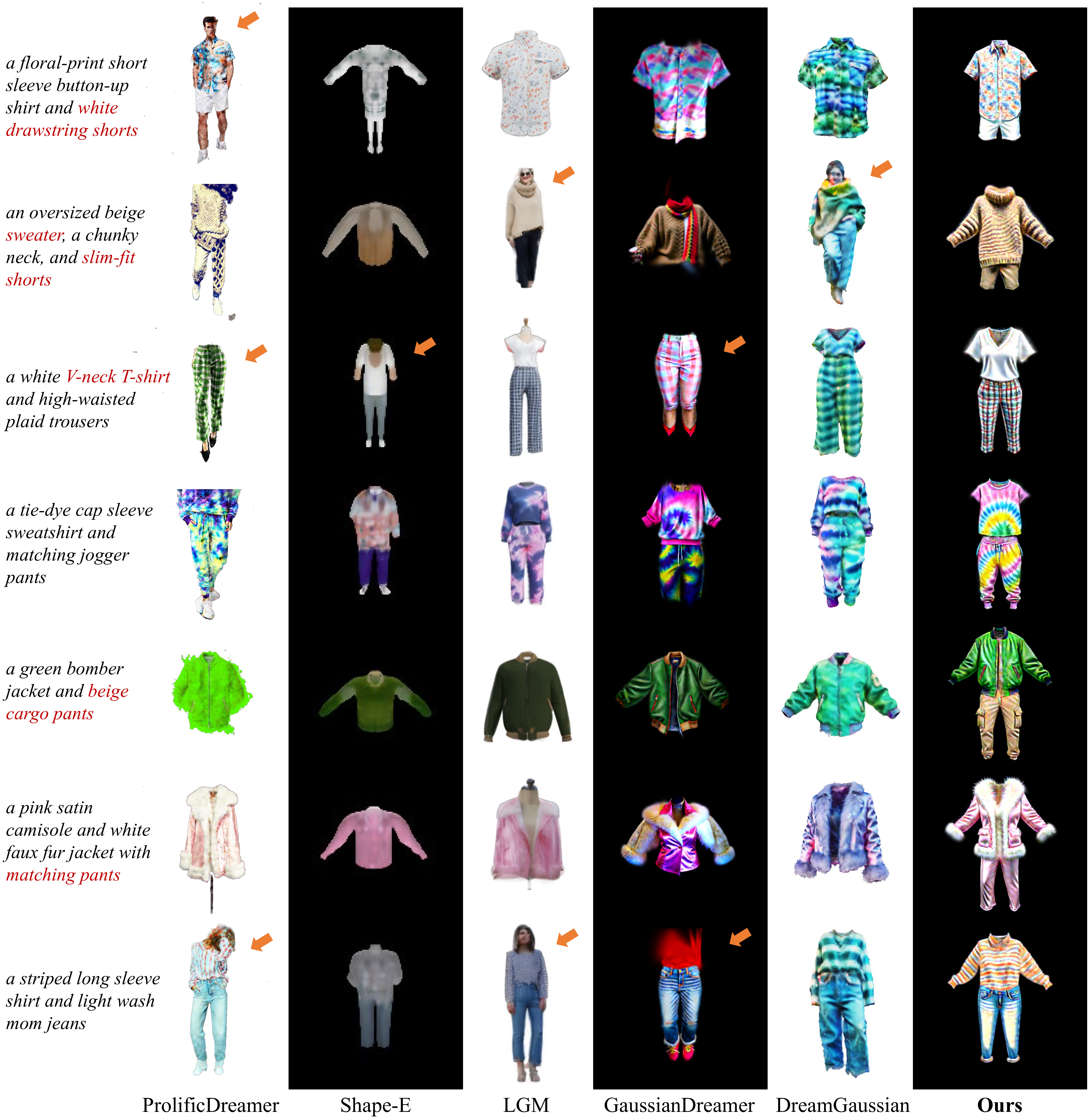}
   \caption{Qualitative comparison of garment generation from text. We compare with recent state-of-the-art 3D generation baselines on seven different garment text descriptions. Note that \textbf{\textcolor{darkred}{red text}} highlights incomplete garment generation, while \textbf{\textcolor{orange}{orange arrows}} indicate geometry artifacts for redundant human body parts.}
   \label{image: quality compare}
   \vspace{-1.0em}
\end{figure*}

\section{Experiments} 
\label{sec:experiment}
\subsection{Implementation details}
\label{subsec: experimental setup}
 
\noindent\textbf{Gaussian Initialization.}
We initialize one 3D Gaussian model with two sets of point clouds. Body part Gaussian has 100k samples on SMPL-X mesh with opacity set to 1.0. Garment sample is counted by Equation~\ref{eq: points}, and the opacity is 0.5. Both colors are represented by spherical harmonics (SH) coefficients~\cite{SH} of degree 0. We use color samples from the UV texture for the body to give a more semantic rendering, and we use neutral grey 0.5 for clothing colors. The whole optimization takes 3000 iterations for both zero-shot and template-guided generation, with the adaptive densifying and pruning from 300 to 800 iterations at an interval of 100 steps. For tight clothes, the prune-only phase is from 1700 to 2100 with an interval of 100 and a scaling factor threshold of 0.007. For loose dress clothes, we only optimize 2500 iterations and prune once at 1800 with 0.5 scaling factor.

\begin{figure}[t]
  \centering
   \includegraphics[width=1.0\linewidth]{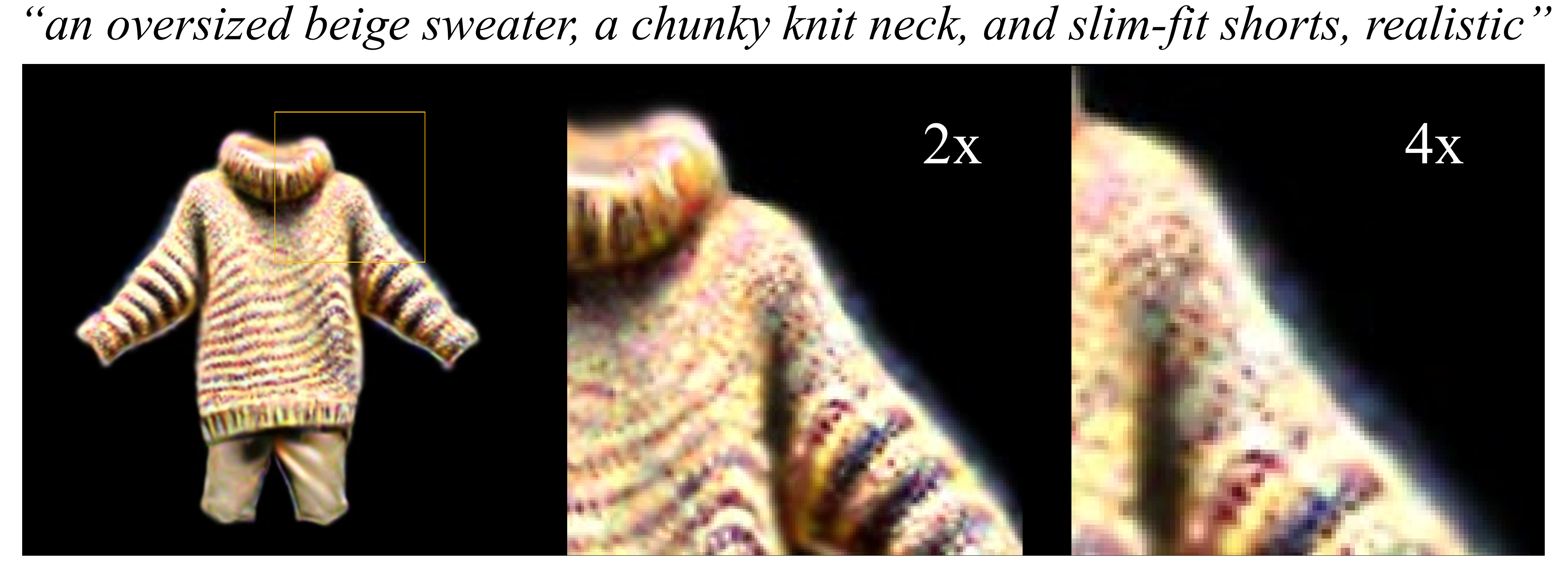}
   \caption{Direct modeling of cloth material. Our method can directly generate complex cloth material without post-processing.}
   \label{image: sweater}
\end{figure}
\begin{figure}[t]
  \centering
   \includegraphics[width=1.0\linewidth]{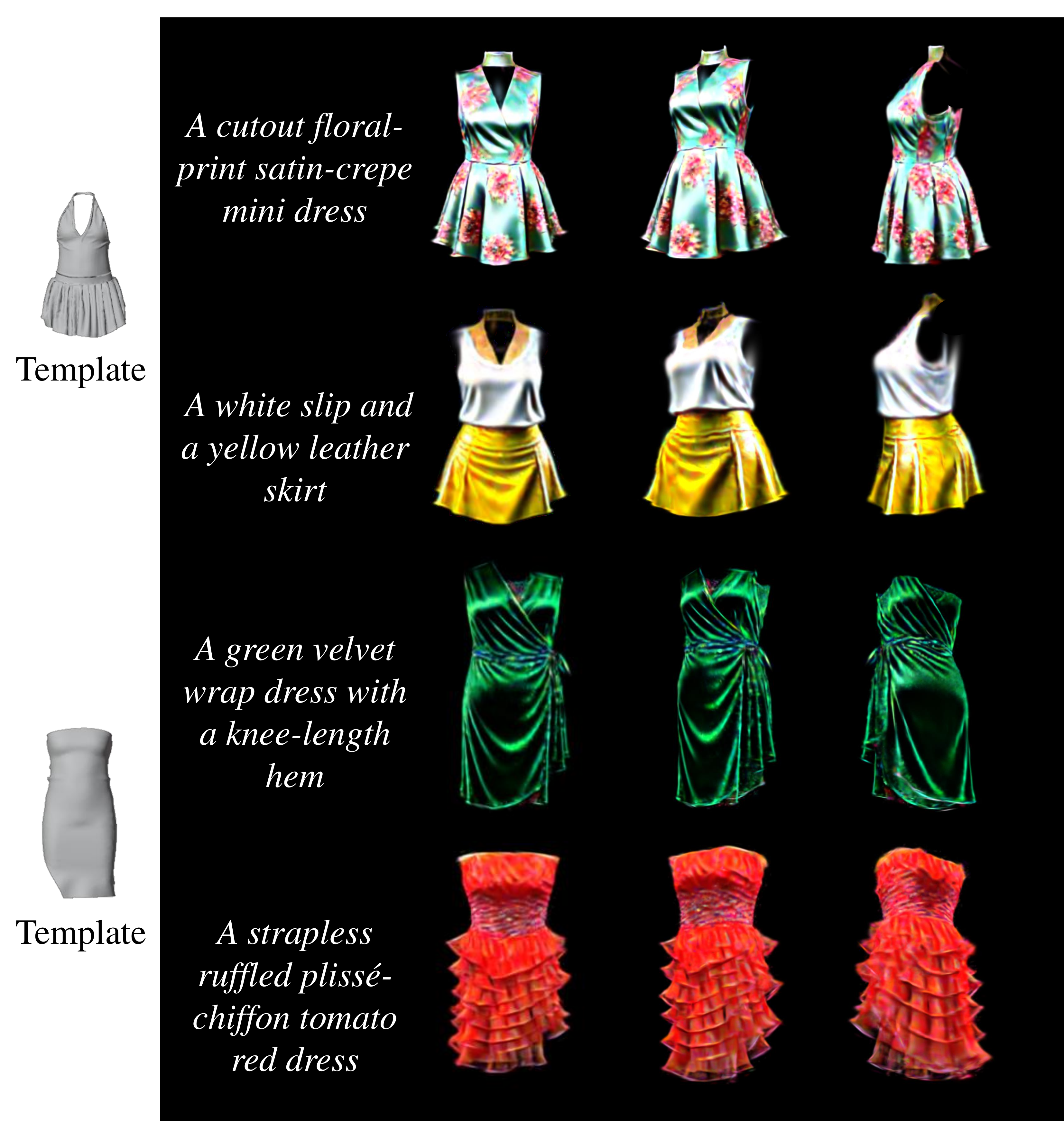}
   \caption{Template-Guided garment generation results.}
   \label{image: template}
\end{figure}

\noindent\textbf{Bidrectional SDS Guidance.}
We use the extended pretrained SD following~\cite{humangaussian} to enhance generation structure by simultaneously denoising RGB and depth. Our framework is implemented using PyTorch based on ThreeStudio~\cite{threestudio}. We use the camera distance range of $[1.5, 2.0]$, fovy range of $[40\degree, 70\degree]$, elevation range of $[-30\degree, 30\degree]$, and azimuth range of $[-180\degree, 180\degree]$. The directional SDS loss weights for RGB and Depth are both set to 0.5. Given a text description, we separate the sentence using "wearing" and define two prompt processors, using the whole sentence to supervise body rendering, and the rest part for clothing. We track the parsed Gaussian render to determine which part is being rendered. The prompt processor is used to calculate SDS loss according to the selected render. The training resolution is 1024, and the entire optimization process takes two hours with batch size 8 on a single NVIDIA A100 (40GB) GPU.

\begin{figure}[t]
  \centering
   \includegraphics[width=1.0\linewidth]{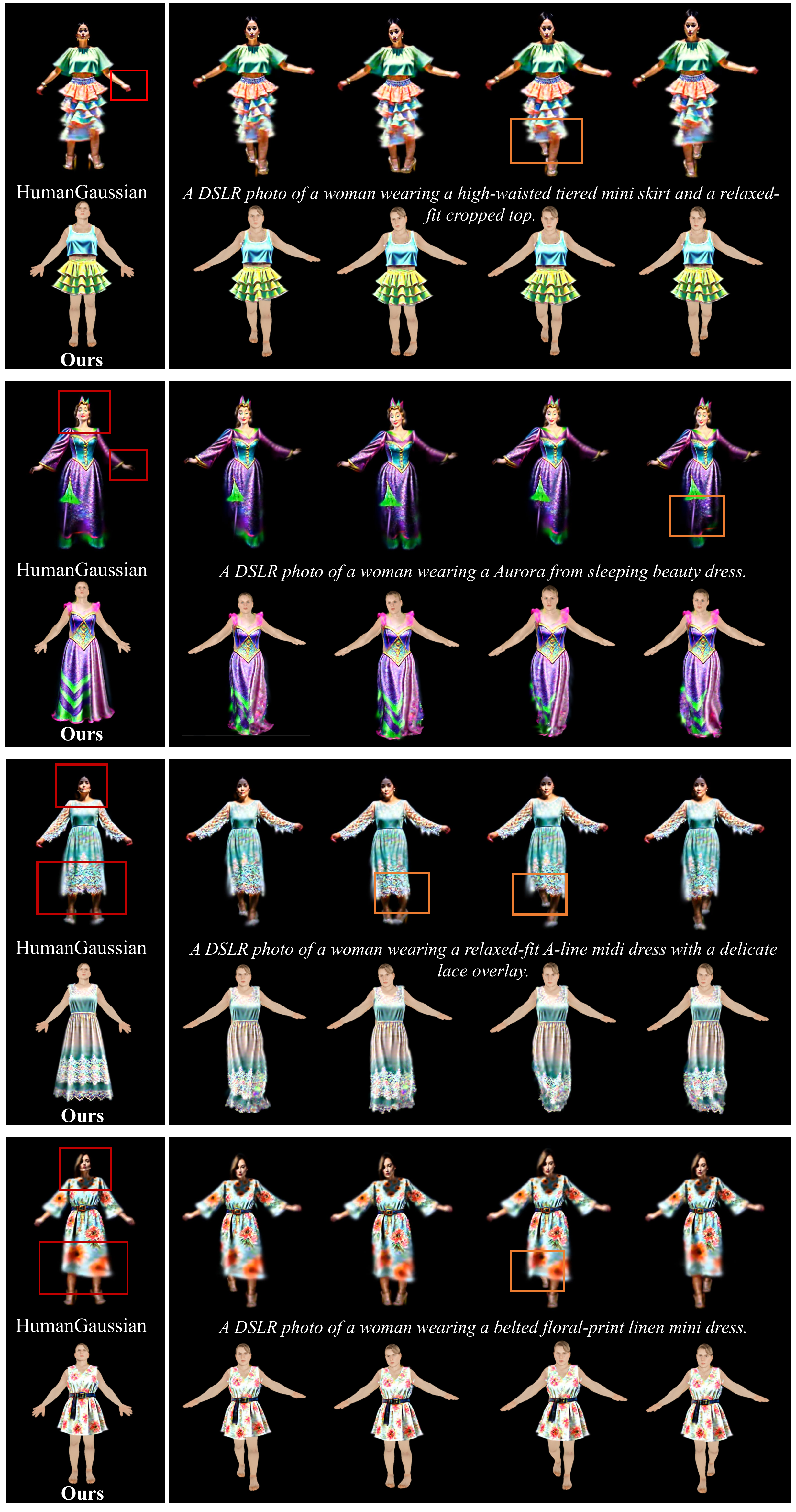}
   \caption{Qualitative comparison of animated clothed avatars with HumanGaussian~\cite{humangaussian}. Note \textbf{\textcolor{darkred}{red boxes}} highlight artifacts for clothed avatar generation, while \textbf{\textcolor{orange}{orange boxes}} indicate garment tearing artifacts during animation. }
   \label{image: compare humangaussian}
   \vspace{-2em}
\end{figure}

\subsection{Qualitative Comparison}
\label{subsec: qualitative comparison}
We compare ClotheDreamer with state-of-the-art text-guided 3D generation methods, including mesh-based Shap-E~\cite{Shap-e}, nerf-based ProlificDreamer~\cite{prolificdreamer}, and Gaussian-based LGM~\cite{lgm}, GaussianDreamer~\cite{gaussiandreamer}, DreamGaussian~\cite{dreamgaussian}. 
We use 100 descriptions generated with GPT-4~\cite{chatgpt} and randomly selected 6 for qualitative comparison. As shown in Figure~\ref{image: quality compare}, our generation achieves the highest overall quality and the finest details in geometry and textures. Note that compared methods tend to give incomplete garments or generate extra human body parts in some cases. 
We also observe that benefiting from 3D Gaussian representation, as shown in Figure~\ref{image: sweater}, our method can generate tricky cloth material like the fluffy appearance of knit yarn.

\begin{table}[t]
  \centering 
  \caption{Quantitative Comparison for clothe generation from text. Our results have the highest text consistency and user preference. }
  \resizebox{0.4\textwidth}{!}{%
  \begin{tabular}{@{}lcc@{}}
    \toprule 
    Method & CLIP Score $\uparrow$ & User Study $\uparrow$ \\
    \midrule
    ProlificDreamer~\cite{prolificdreamer} & 27.319 & 3.35 \\
    Shap-E~\cite{Shap-e}	& 22.475 & 2.46 \\
    LGM~\cite{lgm} & 28.460 & 3.98 \\
    GaussianDreamer~\cite{gaussiandreamer} & 25.406 & 3.38\\
    DreamGaussian~\cite{dreamgaussian} & 25.319 & 3.25\\
    \textbf{ClotheDreamer (Ours)} & \textbf{31.464} & \textbf{4.55} \\
    \bottomrule
  \end{tabular}
  }
  \label{tab: clip score}
\end{table}

\begin{figure}[t]
  \centering
   \includegraphics[width=1.0\linewidth]{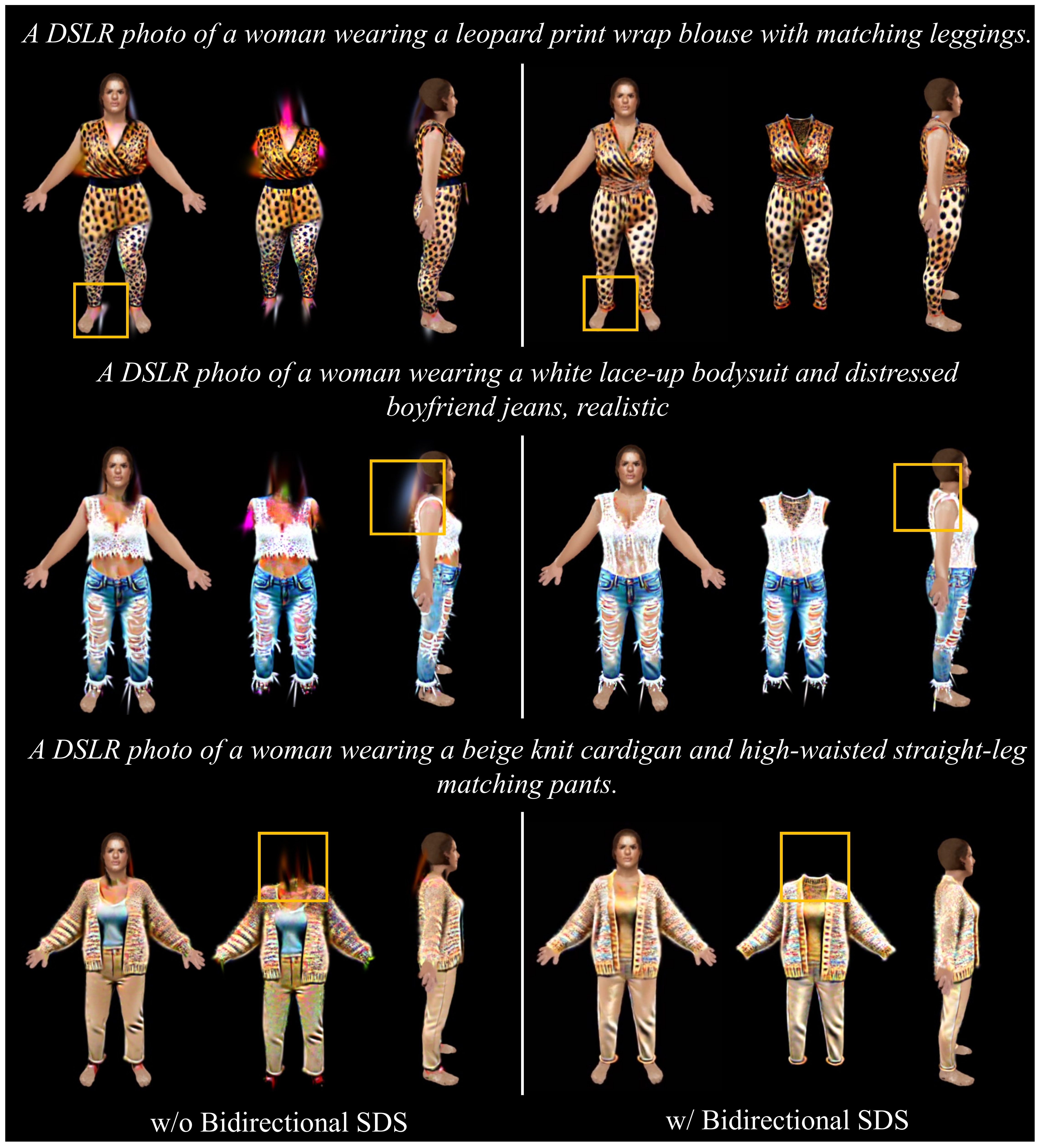}
   \caption{Ablation study for bidirectional SDS Guidance.}
   \label{image: ablation sds}
   \vspace{-2.0em}
\end{figure}

We show some template-guided garment generation results in Figure~\ref{image: template}. The template mesh guides the geometry in general but does not constrain the intricate details. We also compare with HumanGaussian~\cite{humangaussian} for clothed avatar generation and animation in Figure~\ref{image: compare humangaussian}. Integrated with the body, HumanGaussian tends to generate avatars with blurry hands and head shadows on clothes. It also struggles to deal with loose clothes (\eg, short skirts, long dresses). We use two simulated mesh sequences and compare the same motion with HumanGaussian for clothed avatar animation. We randomly select 4 frames from 180. Directly binding the entire avatar on SMPL-X with LBS, HumanGaussian animation can lead to many artifacts, especially in the knee area. 

\begin{figure}[t]
  \centering
   \includegraphics[width=1.0\linewidth]{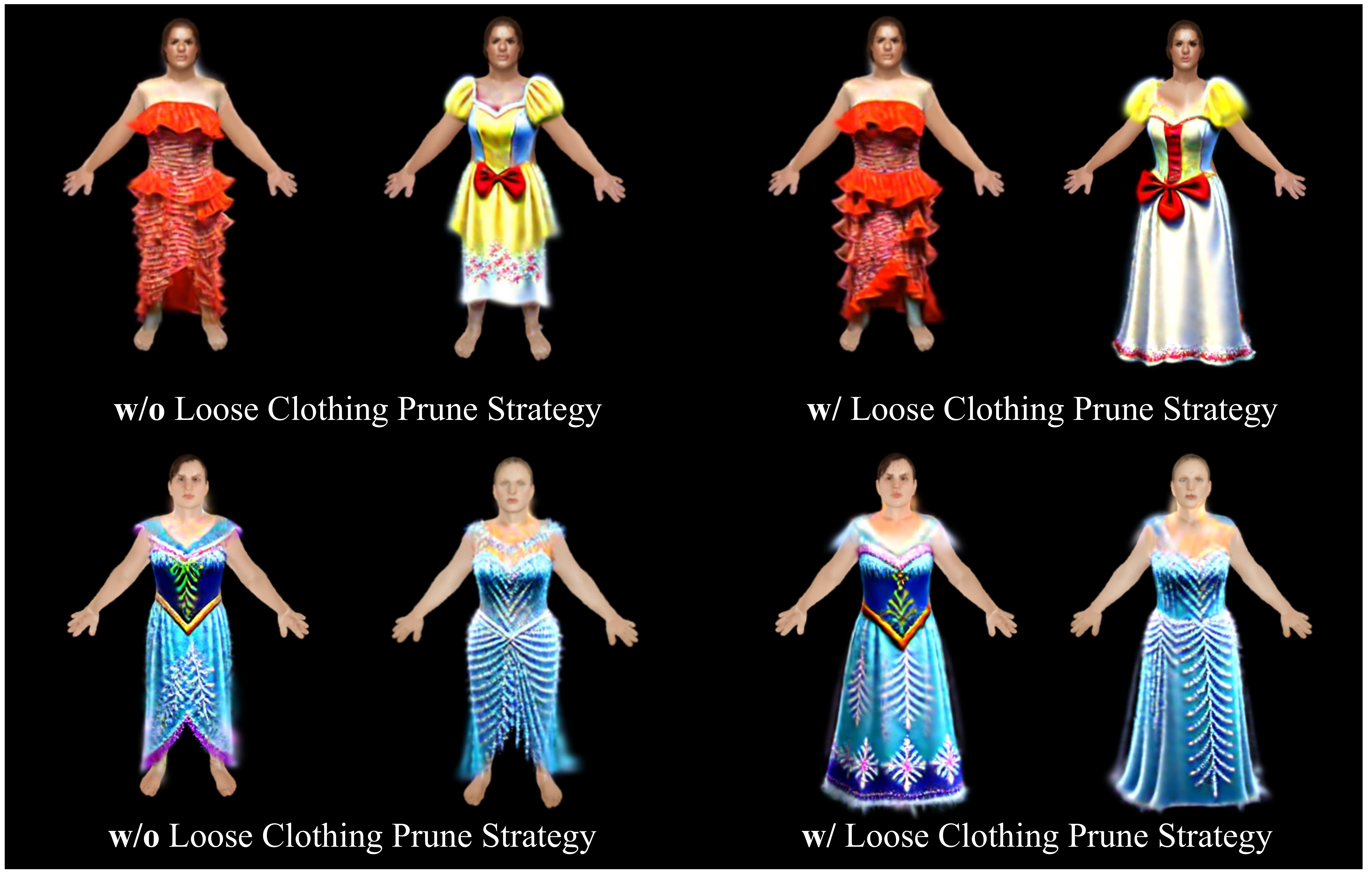}
   \caption{Ablation study of loose clothing prune strategy.}
   \label{image: ablation prune}
\end{figure}

\begin{figure}[t]
  \centering
   \includegraphics[width=1.0\linewidth]{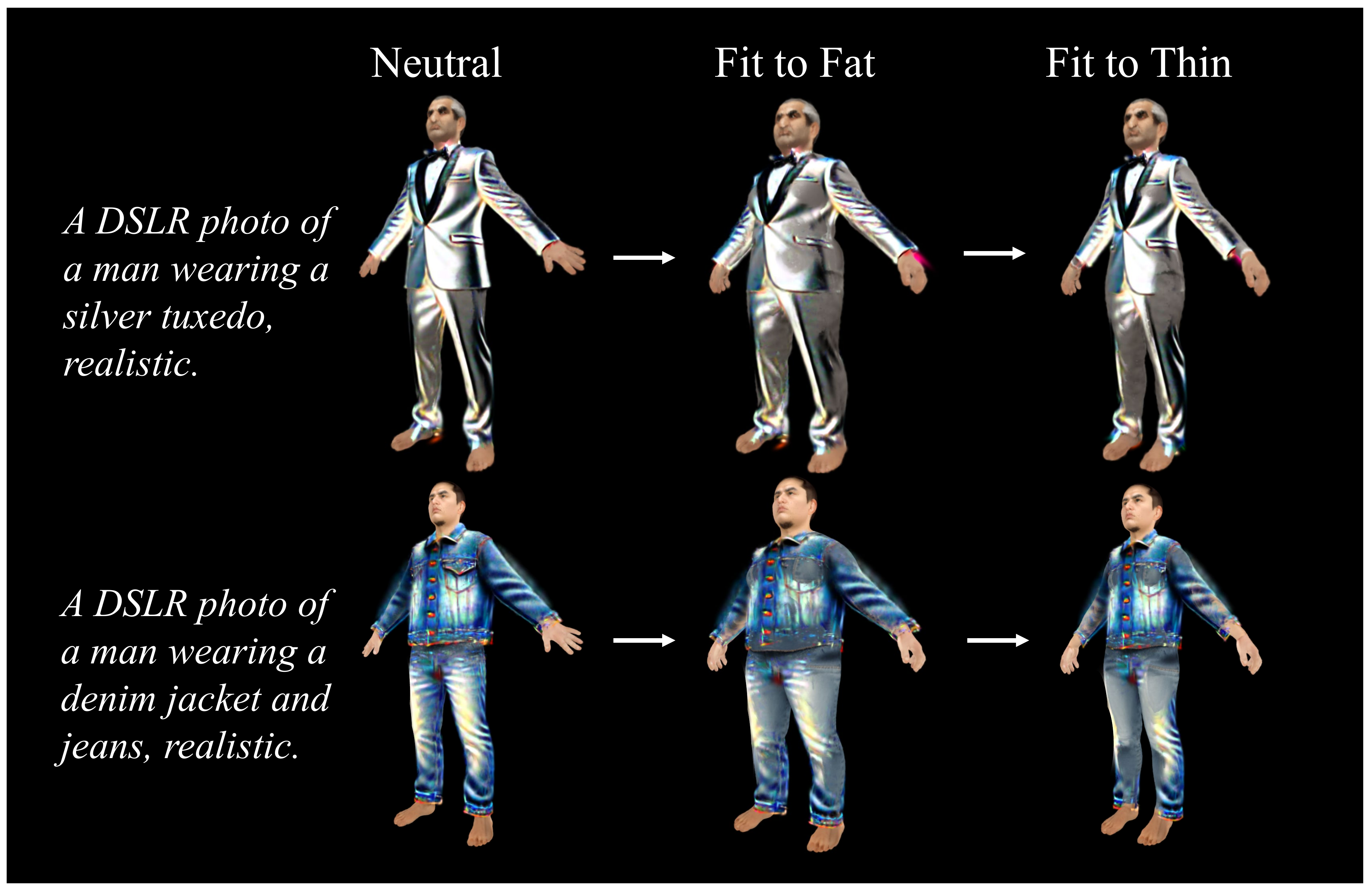}
   \caption{Fit DCGS garment to different body shapes.}
   \label{image: autofitting}
\end{figure}

\subsection{Quantitative Comparison}
\label{subsec: Quantitative comparison}
The rendering of the generated 3D garment from text should closely resemble the input text at the reference view, and demonstrate consistent semantics with the reference under novel views. We evaluate these two aspects with CLIP score~\cite{clip}, which computes the semantic similarity between the novel view and the reference. We generate 100 results for each method with respective training settings. We randomly select 4 frames from each rendered video to calculate the CLIP score. Table~\ref{tab: clip score} shows our results achieve the best text consistency.

\noindent\textbf{User Study.} We further conduct a user study to evaluate generated garments from text. In total, we collected 600 responses for 30 comparison pairs from 20 participants. We ask the participants to score (1-6) based on overall quality and consistency with the given prompt for each method. Table~\ref{tab: clip score} indicates that our method has the highest preference.

\subsection{Ablation Study}
\label{subsec: Ablation Study}

\textbf{Effect of Bidirectional SDS.} In Figure~\ref{image: ablation sds}, we design a variant of our approach by removing the additional SDS guidance on separate clothing renderings. We observe that the generated garments are much worse than our method. Extra floating Gaussian artifacts tend to generate around avatar head area. This aligns with our intuition that additional guidance for clothing individually is needed for better garment generation. In contrast, our bidirectional SDS guidance allows much more clean and robust generation results.

\noindent\textbf{Effect of Loose Clothing Prune Strategy.}  In Figure~\ref{image: ablation prune}, we design a variant of our approach by adopting the pruning strategy used in HumanGaussian~\cite{humangaussian}. Using the scaling factor as a pruning condition proved to be an effective way to eliminate redundant Gaussian splats. However, in loose clothing cases, we observe that this strategy will remove many useful Gaussians by mistake. In comparison, our approach obtains more complete geometry with fine-grained texture details.

\section{Applications}
\label{sec:application}

\noindent\textbf{Auto Fitting.} An important benefit of our approach is that it allows us to decouple the garment generation from the avatar body. Our generated 3D garment asset can be fit to many different avatar body shapes, see Figure~\ref{image: autofitting}. 
Treating mesh as dense point clouds, we can use ICP registration~\cite{icp} to calculate the transformation matrix between two body meshes. Since the generated garment is well-fit on the neutral shape, we can easily bind points on its mesh surface. Based on the body transformation matrix, we can optimize the garment transformation matrix to fit various body shapes, potentially opening doors for a fast virtual try-on.
\section{Conclusion}
\label{sec:application}

We introduce ClotheDreamer, an innovative approach for generating diverse and wearable 3D garments from text prompts. We propose a novel representation, named Disentangled Clothe Gaussian Splatting (DCGS), which can effectively decouple clothing from body. We also propose bidirectional SDS guidance which supervises clothed avatar RGBD rendering separately with pose conditions and a new prune strategy to enhance generation completeness for loose clothes. We further demonstrate our approach versatility by incorporating template mesh primitive for personalized generation. In addition, our DCGS garment can achieve realistic cloth animation with simulated mesh prior. Our experiments and user study indicate that our approach surpasses state-of-the-art methods in terms of appearance and geometry quality.

\noindent\textbf{Limitations and Future work.} While ClotheDreamer shows promising results, it still has several limitations. First, our method currently integrates upper and lower garments, a more refined decoupling would be applicable to more complex try-on scenarios. Second, similar to other SDS-based approaches, in some cases our method also suffers from color oversaturation. We believe that exploring techniques for improving SDS~\cite{sds-1, sds-2} can help mitigate this issue. Lastly, disentangling lighting~\cite{gaussianshader} for 3D Gaussian representation to enhance realism is also an interesting future direction.

\bibliographystyle{eg-alpha-doi} 
\bibliography{egbibsample}       



\end{document}